\documentclass[10pt,twocolumn,letterpaper]{article}

\usepackage{iccv}
\usepackage{times}
\usepackage{epsfig}
\usepackage{graphicx}
\usepackage{amsmath}
\usepackage{amssymb}
\usepackage{comment}
\usepackage{mathtools}
\usepackage{color}
\usepackage{subcaption}
\usepackage{url}            % simple URL typesetting
\usepackage{booktabs}       % professional-quality tables
\usepackage{tabu}           % tabular
\usepackage{multirow}
\usepackage{algorithm}
\usepackage{algorithmicx}
\usepackage{algpseudocode}
\usepackage[dvipsnames]{xcolor}
\usepackage{color}
\usepackage{colortbl}

\definecolor{Gray}{gray}{0.95}

\makeatletter
\@namedef{ver@everyshi.sty}{}
\makeatother
\usepackage{tikz}
\usetikzlibrary{arrows,automata,positioning,shadows}

\usepackage[pagebackref=true,breaklinks=true,colorlinks,bookmarks=false]{hyperref}

\iccvfinalcopy % *** Uncomment this line for the final submission

\def\iccvPaperID{2273} % *** Enter the ICCV Paper ID here

\ificcvfinal\fi
\def\AQA{\mathrm{AQA}}
\DeclareMathOperator{\MSE}{MSE}
\DeclareMathOperator{\concat}{concat}
\def\lefta{\mathrm{left}}
\def\righta{\mathrm{right}}
\usepackage{bm}

\begin{document}

%%%%%%%%% TITLE
\title{Group-aware Contrastive Regression for Action Quality Assessment}

\author{Xumin Yu\thanks{Equal contribution. ~\textsuperscript{\dag}Corresponding author.}, ~Yongming Rao\footnotemark[1], ~Wenliang Zhao, ~Jiwen Lu\textsuperscript{\dag}, ~Jie Zhou\\
Department of Automation, Tsinghua University, China\\
State Key Lab of Intelligent Technologies and Systems, China\\
Beijing National Research Center for Information Science and Technology, China\\
{\tt\small yuxm20@mails.tsinghua.edu.cn; raoyongming95@gmail.com; } \\
{\tt\small zhaowl20@mails.tsinghua.edu.cn; \tt\small \{lujiwen, jzhou\}@tsinghua.edu.cn} \\
}

\maketitle

%%%%%%%%% ABSTRACT
\begin{abstract}
   
Assessing action quality is challenging due to the subtle differences between videos and large variations in scores. Most existing approaches tackle this problem by regressing a quality score from a single video, suffering a lot from the large inter-video score variations. In this paper, we show that the relations among videos can provide important clues for more accurate action quality assessment during both training and inference. Specifically, we reformulate the problem of action quality assessment as regressing the relative scores with reference to another video that has shared attributes (\emph{e.g.}, category and difficulty), instead of learning unreferenced scores. Following this formulation, we propose a new \textbf{Co}ntrastive \textbf{Re}gression (CoRe) framework to learn the relative scores by pair-wise comparison, which highlights the differences between videos and guides the models to learn the key hints for assessment. In order to further exploit the relative information between two videos, we devise a group-aware regression tree to convert the conventional score regression into two easier sub-problems: coarse-to-fine classification and regression in small intervals. To demonstrate the effectiveness of CoRe, we conduct extensive experiments on three mainstream AQA datasets including AQA-7, MTL-AQA and JIGSAWS. Our approach outperforms previous methods by a large margin and establishes new state-of-the-art on all three benchmarks.
 
\end{abstract}

%%%%%%%%% BODY TEXT
\section{Introduction}
  Action quality assessment (AQA), which aims to evaluate how well a specific action is performed, has attracted growing attention in recent years since it plays a crucial role in many real world applications including sports~\cite{firstaqa,DBLP:conf/wacv/ParmarM19,DBLP:conf/icvs/JugPDK03,asdasfaf,DBLP:conf/eccv/PirsiavashVT14,DBLP:journals/corr/abs-1904-04346,DBLP:conf/bmvc/VenkataramanVT15,DBLP:conf/cvpr/ParmarM17}, healthcare~\cite{DBLP:conf/ipcai/MalpaniVCH14,DBLP:journals/pami/ZhangL15,Sharma2014VideoBA,10.1145/2072545.2072550,DBLP:conf/miccai/ZiaSBSCE15,DBLP:journals/cars/ZiaSBSE18} and others~\cite{DBLP:journals/corr/DoughtyDM17, DBLP:conf/cvpr/DoughtyMD19}.  Unlike conventional action recognition tasks that focus on action classification~\cite{DBLP:conf/icml/JiXYY10,DBLP:conf/eccv/WangXW0LTG16,DBLP:conf/cvpr/Wang0T15,DBLP:conf/nips/SimonyanZ14,DBLP:conf/mmm/LiCHYCX19,DBLP:conf/iccv/Feichtenhofer0M19,DBLP:conf/cvpr/0004GGH18} and detection ~\cite{DBLP:conf/iccv/ZhaoXWWTL17,DBLP:journals/corr/abs-1907-09702,DBLP:conf/cvpr/TangDRZZZL019,DBLP:conf/cvpr/YeungRMF16,DBLP:journals/corr/MontesSN16}, AQA is more challenging as it requires the model to predict fine-grained scores from videos that describe the same action. 
  % We argue that the core of this task is comparison, because judges determine the scores of an action based on the comparison between what they have seen and what they want to see.
  Considering the differences between videos and large variations in scores, we argue that a key to addressing this problem is to discover the differences among the videos and predict scores based on the differences. 
  
  \begin{figure}
  \centering
  \includegraphics[width=\linewidth]{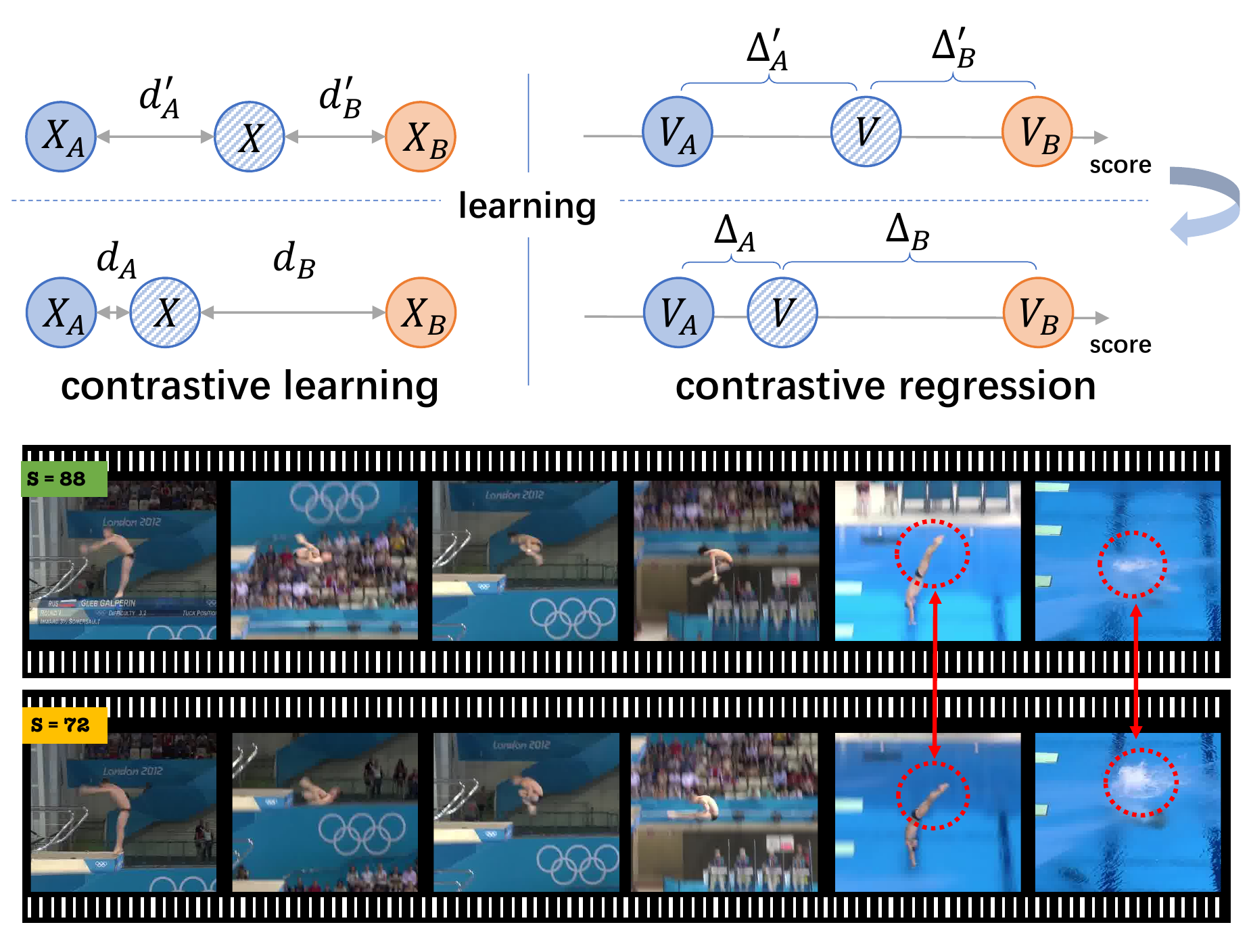}
  \caption{Our Contrastive Regression (\textit{CoRe}) framework for action quality assessment. Inspired by contrastive learning that learns representation by encouraging the distances of samples (\eg, $d_A$ and $d_B$) to reflect their semantic relationship, we learn an AQA model to regress the relative scores (\eg, $\Delta_A$ and $\Delta_B$) to reflect the differences of action quality among videos. By comparing two videos with different scores, \textit{CoRe} encourage the model to learn from differences between videos for assessment.
  }
  \label{fig:insight}
  \vspace{-10pt}
\end{figure}

  Many efforts have been made to tackle this problem over the past few years~\cite{DBLP:conf/iccv/JiaHuiaction,DBLP:journals/corr/abs-1904-04346,DBLP:conf/cvpr/DoughtyMD19,score_figure_skating,DBLP:conf/wacv/ParmarM19}. Most of them formulate the AQA as a regression problem, where the scores are directly predicted from a single video. While some promising results have been achieved, AQA still faces three challenges. First, since the score labels are usually annotated by human judges (\emph{e.g.}, the score of the diving game is calculated by summarizing scores from different judges, then multiplied by the degree of difficulty), the subjective appraisal of judges makes accurate score prediction quite difficult. Second, the difference between videos for AQA is very subtle, since actors are usually performing the same action in a similar environment. Last, most current models are evaluated based on Spearman's Rank, which may not faithfully reflect the prediction performance (see our discussions in Section~\ref{sec:setup}).
  
  Towards a better AQA framework that can utilize the differences among the videos to predict the final rating, we borrow the merits from the concept of contrastive learning~\cite{he2020momentum,chen2020simple}. Contrastive learning (Figure~\ref{fig:insight}, top-left) aims to learn a better representation space where the distance $d_A$ between two similar samples $X,X_A$ is enforced to be small while the distance $d_B$ between the dissimilar ones $X,X_B$ is encouraged to be large. Therefore, the distance in the representation space can already reflect the semantic relationship between two samples (\ie, if they are from the same category). Analogically, in the context of AQA, we aim to learn a model that can map the input video into the score space where the differences between the action qualities can be measured by the relative scores ($\Delta_A,\Delta_B$). Motivated by this, we propose a \textbf{Co}ntrastive \textbf{Re}gression (CoRe) framework for the AQA task. Unlike previous works which aim to predict the scores directly, we propose to regress the relative scores between an input video and several exemplar videos as references.

%   Human judges assess quality score of an action from two clues: one is how well the current action is performed and the other is their experiences. They may consider the difference between the current action and the prototype action in their minds, then give a score based on the comparison, which also leads to the subjective preferences of different judges~\cite{musdl}. Inspired by the assessment process of human judges, we propose to reformulate the AQA problem as regressing relative scores with reference to another video that has shared attributes (\emph{e.g.}, performing the same action or having the same degree of difficulty of the action), instead of learning unreferenced scores. 
  %is explicitly guided by known scores given by the human judges and encouraged to predict scores based on the subtle differences between the current video and the exemplar.
  %Inspired by the contrastive learning strategy in metric learning literature~\cite{hadsell2006dimensionality}, 
%   We propose a new \textbf{Co}ntrastive \textbf{Re}gression (\textit{CoRe}) framework to learn the relative scores based on the exemplars. By introducing an exemplar for score prediction, the regressor is explicitly guided by known scores given by the human judges and encouraged to predict scores based on the subtle differences between the current action and the exemplar. 
  
  Moreover, as a step towards more accurate score prediction, we devise a group-aware regression tree (GART) to convert the relative score regression into two easier sub-problems: (1) coarse-to-fine classification. We first divide the range of the relative score into several non-overlapping intervals (\ie, groups) and then use a binary tree to allocate the relative score to a certain group by performing classification progressively;  (2) regression in a small interval. We perform regression inside the group where the relative score lies and predict the final score. 
%   There are mainly two strengths of our method. 
  % Firstly, contrasting the input video with another exemplar video help the model to focus on some discriminative regions in the video that are crucial to acquire the score instead of some less informative ones such as the background. Secondly, the GART formulate the whole regression procedure as a sequence of classification problems and an easier regression problem defined on a small interval, which can perform the regression in a coarse-to-fine manner and can provide results with better interpretability.
%   and regression in small intervals.
%   The strengths of our method are: 1) the model predicts the score of an action in a manner similar to human judges, which also reduce the effects of the subjective appraisal in single score labels by highlighting the difference between two action-score pairs; 2) by comparing different actions, the model is guided to pay more attention to the details that actually affect the final score, instead of the background environment or other confounders.
  As another contribution, we design a new metric, called relative L2-distance (R-$\ell_2$) to more precisely measure the performance of action quality assessment by considering the intra-class variance.
  
  To verify the effectiveness of our method, we conduct extensive experiments on three mainstream AQA datasets containing both Olympic and surgical actions, namely AQA-7~\cite{DBLP:conf/wacv/ParmarM19}, MTL-AQA~\cite{DBLP:journals/corr/abs-1904-04346} and JIGSAWS~\cite{gao2014jhu}. Experiments results demonstrate our method largely outperforms the state-of-the-art on the three benchmarks under the Spearman’s Rank Correlation (81.0\% to 84.0\% on AQA-7, 92.7\% to 95.1\% on MTL-AQA and 70\% to 85\% on JIGSAWS) and new proposed R-$\ell_2$ metric, which clearly shows the advantages of our proposed contrastive regression framework.

\section{Related Work}

\begin{figure*}
  \centering
  \includegraphics[width = 0.9\linewidth]{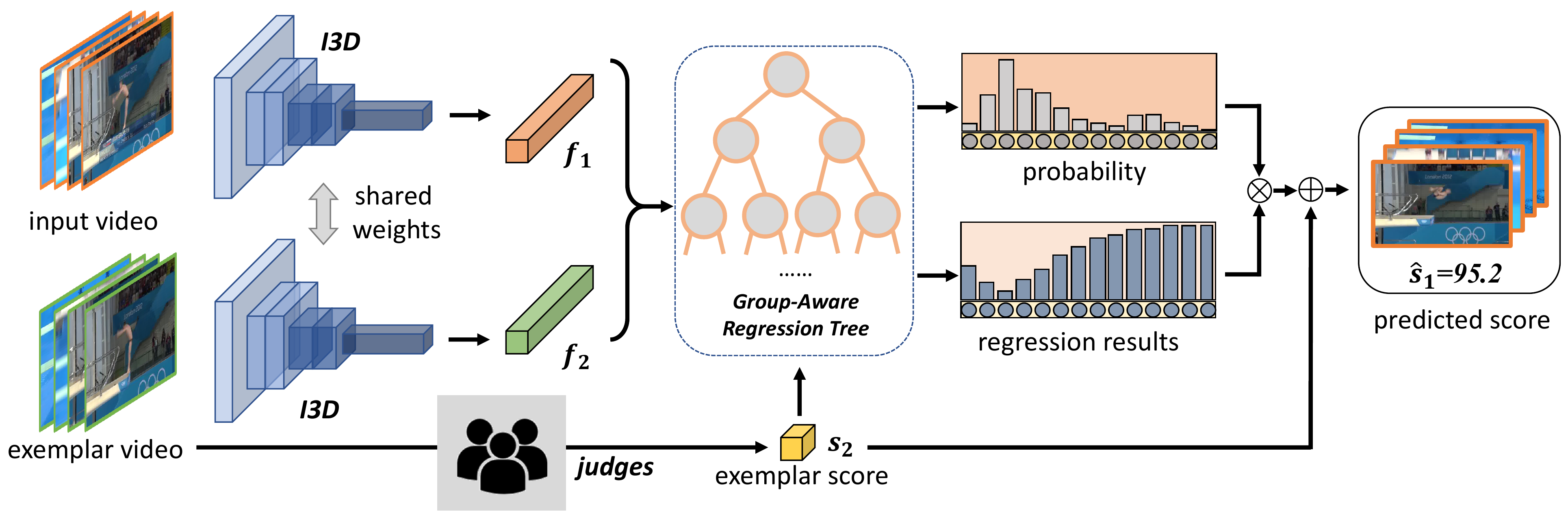}
  \vspace{-5pt}
  \caption{The pipeline of our proposed group-aware contrastive regression method. We first sample an exemplar video for each input video according to the category and degree of difficulty of the action. We then feed the video pair into a shared I3D backbone to extract spatio-temporal features and combine these two features with the reference score of the exemplar video. Finally, we pass the combined feature to the group-aware regression tree and obtain the score difference between the two videos. During inference, the final score can be computed by averaging the results from multiple different exemplars.}
  \label{fig:pipeline}
  \vspace{-5pt}
\end{figure*}

% \noindent \textbf{Action Quality Assessment:}
The past few years have witnessed the rapid development of AQA. 
% Many methods are proposed to improve the performance of AQA. 
The mainstreams of AQA methods formulate AQA as a regression task based on reliable scores labels given by expert judges. For example, Gordan \etal~\cite{firstaqa} propose to use the trajectory of the skeleton to solve the problem of gymnastic vault action quality assessment in their pioneer work. Pirsiavash \emph{et al.}~\cite{DBLP:conf/eccv/PirsiavashVT14} use DCT to encode body pose as input features. SVR~\cite{Basak2007SupportVR} is also used to build the mapping from the features to the final score. 
% Venkataraman \emph{et al.}~\cite{DBLP:conf/bmvc/VenkataramanVT15} extend~ \cite{DBLP:conf/eccv/PirsiavashVT14} by replacing the DCT with approximated entropy to encode the poses. which improves the performance of the previous works by nearly 4\%. 
Thanks to the great success of deep learning in action recognition tasks, Parmar \emph{et al.}~\cite{DBLP:conf/cvpr/ParmarM17} show that the spatio-temporal features from C3D ~\cite{DBLP:conf/iccv/TranBFTP15} can better encode the video data and significantly improve the performance. They also propose a large-scale AQA dataset and explore all-action models to further enhance the scoring performance. Following~\cite{DBLP:conf/cvpr/ParmarM17}, Xu~\etal~\cite{score_figure_skating} propose a model containing two LSTM to learn the multi-scale features of videos. Pan~\etal~\cite{DBLP:conf/iccv/JiaHuiaction} propose to use spatial and temporal relation graphs to model the interaction among the joints. In addition, they also propose to use I3D~\cite{DBLP:conf/cvpr/CarreiraZ17} as a stronger backbone network to extract spatio-temporal features. Parmar~\etal~\cite{DBLP:journals/corr/abs-1904-04346} propose a larger AQA dataset with more annotations for various tasks. The idea of multi-task learning is also introduced to improve the model capacity for AQA. Recently, Tang~\etal~\cite{musdl} propose a new uncertainty-aware score distribution learning (USDL) to reduce the underlying ambiguity of the action score labels from human judges. Different from this line of works, several methods~\cite{DBLP:journals/pami/ZhangL15, DBLP:journals/corr/DoughtyDM17, DBLP:conf/cvpr/DoughtyMD19, DBLP:conf/iccv/BertasiusPYS17a} formulate AQA as a pair-wise ranking task. However, they mainly focus on longer and more ambiguous tasks and only predict an overall ranking, which might limit the application of AQA where some quantitative comparisons are required. In this work, we present a new contrastive regression framework to simultaneously rank videos and predict accurate scores, which makes our method distinguished from previous works.

\section{Approach}

\label{approach}

%In this part, we will introduce our Group-aware contrastive regression approach for action %quality assessment.
% In general, our approach is composed of two steps.
%The first step is to construct the input-exemplar pair and construct the combined feature after extracting features of each video.
% The second step is to regress the combined feature to relative score by a group-aware regression tree.
%Each leaf of this tree represent a certain group gathering similar pair. %Regression will be carried out within the most possible group and a precise %difference value will be predicted.
% In this section, we will introduce the group-aware contrastive regression method. In Section~\ref{sec:p1}, we first detail the proposed contrastive regression framework. Then, we will show how to employ the framework to achieve accurate score prediction with a group-aware regression tree network in Section~\ref{sec:p2}. 
The overall framework of our method is illustrated in Figure~\ref{fig:pipeline}. We will describe our method in detail as follows.

\subsection{Contrastive Regression}
\label{sec:p1}
\paragraph{Problem Formulation.} 
%first define the problem we want to solve
Most existing works~\cite{DBLP:conf/iccv/JiaHuiaction,DBLP:journals/corr/abs-1904-04346,DBLP:conf/cvpr/DoughtyMD19,score_figure_skating,DBLP:conf/wacv/ParmarM19,musdl} formulate AQA as a regression task, where the input is a video containing the target action and the output is the predicted quality score of the action. Note that in some AQA tasks (\eg, diving), each video is associated with a degree of difficulty for each video (which is a known constant). The final score is the multiplication of the action quality score (\ie, raw score) and the degree of difficulty. Since the degree of difficulty is already known, we only need to predict the action quality score following~\cite{musdl}.  Formally, given the input video $v = \{F^i\}_{i=1}^L$ with action quality label $s$,
the regression problem is to predict the action quality $\hat{s}$ based on the input video:
\begin{small}
\begin{equation}%\small
    \hat{s} = \mathcal{R}_{\Theta}(\mathcal{F}_{\mathcal{W}}(v)),
\end{equation}
\end{small}
where $\mathcal{R}_{\Theta}$ and $\mathcal{F}_{\mathcal{W}}$ are the regressor model and the feature extractor parameterized by $\Theta$ and $\mathcal{W}$, respectively. The regression problem is usually solved by minimize the mean-square error between the predicted score and the ground-truth score:
\begin{small}
\begin{equation}%\small
    \mathcal{L}_{\AQA}(\Theta,\mathcal{W} |v) = \MSE(\hat{s}, s),
\end{equation}\end{small}
where $\Theta$ and $\mathcal{W}$ are the parameters of regression model and feature extractor.

However, since the action videos are usually captured in similar environments (\eg, diving competitions often take place in aquatics centers), it is difficult for the model to learn the diverse scores based on videos with subtle differences. To this end, we propose to reformulate the problem as regressing relative score between the input and an exemplar. Let $v_m = \{F^i_m\}_{i=1}^{L_m}$ denotes the input video, and $v_n = \{F^i_n\}_{i=1}^{L_n} $ denotes the exemplar video with score label $s_n$. The regression problem can be re-written as:
\begin{small}\begin{equation}%\small
    \hat{s}_m = \mathcal{R}_{\Theta}(\mathcal{F}_{\mathcal{W}}(v_m), \mathcal{F}_{\mathcal{W}}(v_n)) + s_n.
\end{equation}\end{small}
This formulation can be also viewed as a form of residual learning~\cite{he2016deep}, where we aim to regress the difference of the scores between the input video and a reference video.

% and contrasting learning in embedding learning~\cite{hadsell2006dimensionality}, which can be viewed as a residual learning method among different videos or a contrastive learning method for regression.  

% In the following sections, we will describe how to implement the framework for AQA problem, including the strategy to construct the video pairs, the detailed network architecture and the learning algorithm. 

\paragraph{Exemplar-Based Score Regression.} 
%first brief why we want to construct a pair for regression. then show how to construct dataset, then show how to construct features.
We now describe how to implement the CoRe framework for the AQA problem. Since we aim to regress the relative score between the input and the exemplar, how to select the exemplar becomes critical. To make the input and the exemplar comparable, we tend to select the video that shares some certain attributes (\eg, category and degree of difficulty) with the input video as the exemplar.
% how to select exemplars becomes an important problem that needs to be resolved. Our solution is to choose the videos that are the most similar to the current video. 
% We adopt an exemplar-based regression strategy to reduce the affect from human judges' subjective bias. We want the network to imitate human judges to complete this comparison process. So, how do a human judge choose an exemplar for a current action? It can be the same class action or the different action by the same performer. 
% Limited by the annotations in current AQA dataset, we choose an exemplar for an input video from the videos with the shared attributes (\emph{e.g.}, category and degree of difficulty).
Formally, given an input video $v_m$ and the corresponding exemplar $v_n$, we first use an I3D~\cite{DBLP:conf/cvpr/CarreiraZ17} to extract the features $\{f_n,f_m\}$ following~\cite{musdl,DBLP:journals/corr/abs-1904-04346}, and then aggregate them with the score of the exemplar $s_n$:
\begin{small}\begin{equation}
    f_{(n,m)} = \concat([f_n,f_m,s_n/\epsilon]),
\end{equation}\end{small}
% we sample a video $v_n$ with the label $s_n$ from the training subset as the exemplar. These two videos construct a pair $S_{pair}^{m,n} = \{v_m ,v_n ,s_n\}$. 
% TODO: move to the supplementary
% To extract video features, we follow the practice in ~\cite{musdl,DBLP:journals/corr/abs-1904-04346} by utilizing a sliding window to segment each video into $X$ overlapping snippets with a fixed $Y$ consecutive frames and further sending those snippets into a backbone of Inflated 3D ConvNets (I3D)~\cite{DBLP:conf/cvpr/CarreiraZ17} followed by a temporal average pooling layer, resulting in video-level feature for input video and its exemplar as $\{f_n,f_m\}$. 
% 
% After encoding the current video and the exemplar video, we combine them into a totally new combined feature:
% \begin{equation}\small
% \end{equation}
where $\epsilon$ is a normalizing constant to make sure ${s_n}/{\epsilon}\in [0, 1]$. We then predict the score difference of the pair through a regressor $ \mathcal{R}_{\Theta}$ as $\Delta s =  \mathcal{R}_{\Theta}(f_{(n,m)})$.

\subsection{Group-Aware Regression Tree}
\label{sec:p2}

\begin{figure}
  \centering
  \includegraphics[width=\linewidth]{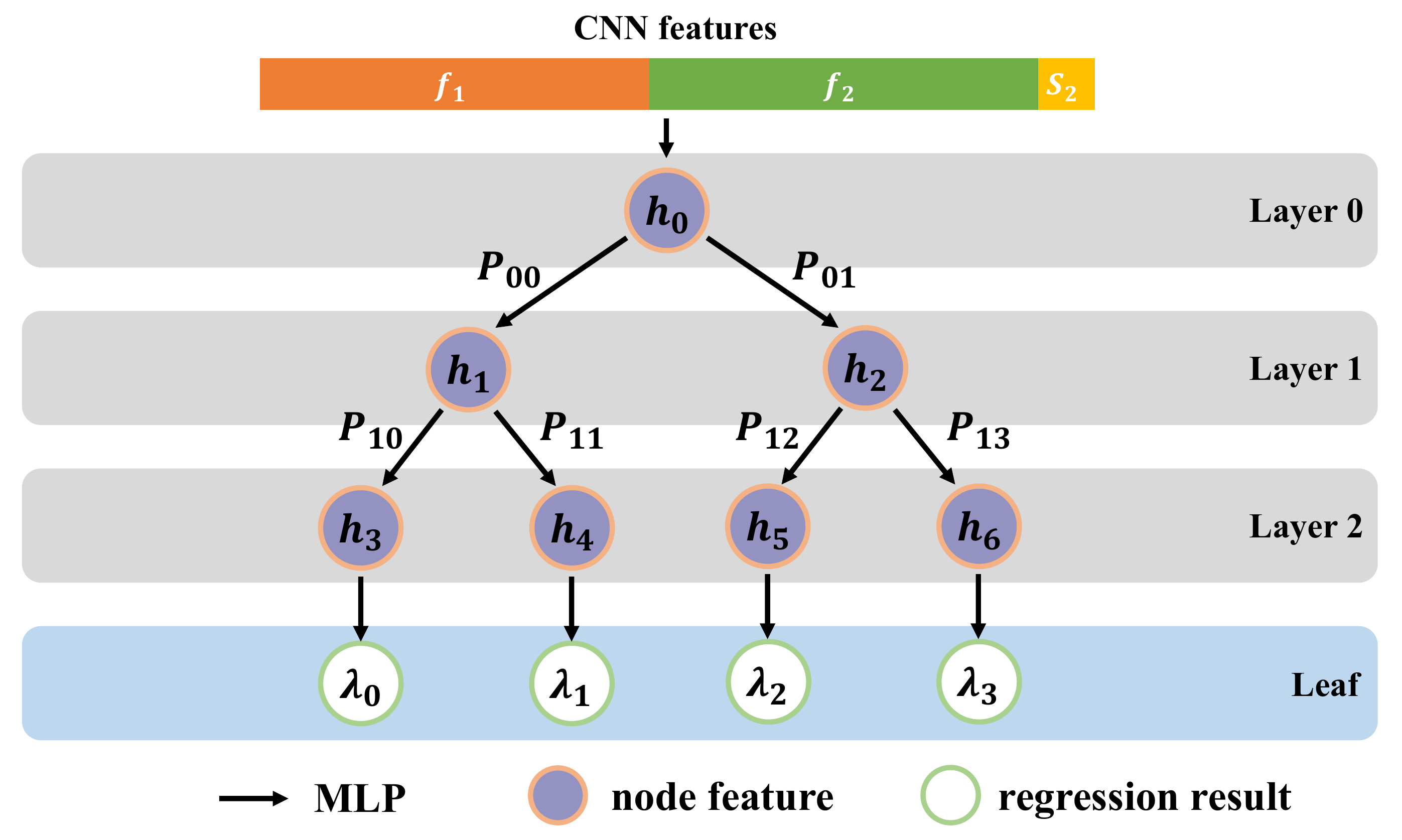} \vspace{-5pt}
  \caption{The architecture of the proposed group-aware regression tree. Given the video features and the reference score, the regression tree determines the score difference in a coarse-to-fine manner, where a sequence of binary classification tasks is performed at first (purple nodes) and the regression modules in the leaf layer then give the final prediction (white nodes). 
  }
  \label{fig:tree}
  \vspace{-5pt}
\end{figure}

% Humans usually compare two actions by deciding which one is better at first, then figuring out how much this one is better. The progressive strategy brings more accurate results and better explanation ability.

Although the contrastive regression framework can predict the relative score $\Delta s$, $\Delta s$ usually takes values from a wide range (\eg, for diving, $\Delta s\in [-30, 30]$). Therefore, predicting $\Delta s$ directly is still of great difficulty. To this end, we devise a group-aware regression tree (GART) to solve the problem in a divide-and-conquer manner. Specifically, we first divide the range of $\Delta s$ into $2^d$ non-overlapping intervals (namely ``groups''). We then construct a binary regression tree with $d-1$ layers, of which the leaves represent the $2^d$ groups, as is illustrated in Figure~\ref{fig:tree}.
% Inspired by this, we design a novel group-aware regression tree to map the combined feature to the relative score, which divides the regression task into several binary classification problems in each layer and an easier regression problem in much smaller intervals. The architecture of the proposed regression tree is  To improve the regression accuracy, we use each node in the leaf layer to represent a small interval (called \textit{group}) instead of a single value. 
% There is no overlap between any two groups and the parent node in the tree represents a bigger group contains all his children nodes. We carefully separate the score interval to avoid introducing more problems from data imbalance. In fact, gathering the data with adjacent label together, espeacially in assessment task~\cite{AgeGroup}, is helpful to further reduce the underlying ambiguity in score label provided by human judges.
The decision process of group-aware regression tree follows a coarse-to-fine manner: in the first layer, we determine whether the input video is better or worse than the exemplar video; in the following layers, we gradually make a more accurate prediction about how much the input video is better/worse than the exemplar. Once we have reached the leaf nodes, we can know which group the input video should be classified and we can then perform regression in the corresponding small interval.

% In the leaf layer, the regression modules give the final prediction. This design actually imitates the humans' behavior in score prediction tasks.

\vspace{2pt}
\noindent \textbf{Tree Architecture. } 
 We adopt the binary tree architecture to perform the regression task. To begin with, we perform an MLP to $f_{(n, m)}$ and use the output as an initialization of the root node feature. We then perform the regression in a top-down manner. Each node takes the output feature from its parent node as input and produces the binary probability together with the updated feature. The probability of each leaf node can be computed by multiplying all the probabilities along the path to the root. We use the Sigmoid to map the output of each leaf node to $[0, 1]$, which is the predicted score difference w.r.t. the corresponding group. 
 
\begin{figure}
    \centering
    \includegraphics[width=\linewidth]{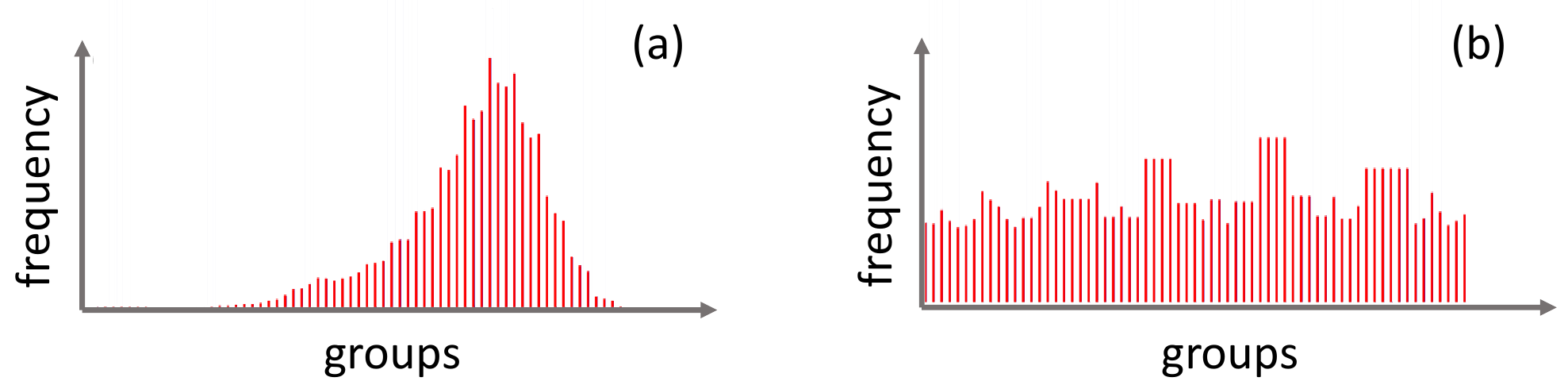}  \vspace{-10pt}
    \caption{The distribution of the differences of scores in the training set under different partition strategy. (a) Uniform partition. We can observe a large variation of frequency among different groups. (b) The proposed grouping strategy in Equation~\eqref{equ:grouping}. The training pairs belonging to each group are balanced.} \vspace{-5pt}
    \label{fig:grouping_strategy}
\end{figure}

 We then describe our partition strategy to define the boundary of each group.
First, we collect the list of score differences of all possible training video pairs $\bm{\delta} = [\delta_1, ... , \delta_T]$. Then, we sort the list in an ascending order to obtain $\bm{\delta}^* = [\delta_1^*, ... , \delta_T^* ]$. Given the group number $R$, the partitioning algorithm gives the bounds of each interval $\mathcal{I}^r = (\zeta_{\lefta}^r,\zeta_{\righta}^r)$ as:
\begin{small}\begin{equation}
\begin{aligned}
    &\zeta_{\lefta}^r = \bm{\delta}^*\left(\left\lfloor (T-1) \times  \frac{ (r-1)}{R}\right\rfloor\right),\\    
    &\zeta_{\righta}^r = \bm{\delta}^*\left(\left\lfloor (T-1) \times  \frac{ r}{R}\right\rfloor]\right), \forall i=1,2,\ldots,R,\label{equ:grouping}
\end{aligned}
\end{equation}\end{small}
where we use $\bm{\delta}^*(i)$ to represent the $i-$th element of $\bm{\delta}$. It is worth noting that the partition strategy is non-trivial. If we simply uniformly divide the whole range into multiple groups, the pairs of videos in the training set of which the differences of scores lie in some certain group may be unbalanced (see Figure~\ref{fig:grouping_strategy} for details).

% %写一个algorithm
% \begin{algorithm}[t]
% \caption{Interval Partitioning Algorithm for Leaf nodes}
% \label{alg:A}
% \begin{algorithmic}[1]
% \Require Depth of the regression tree $d$, training set $D$;
% \Ensure Regression interval of each leaf $\{\mathcal{I}^r\}_{r=1}^R$
% \State Number of Leaf $R = 2^d$;
% \State Collect score differences $\Delta = [\delta_1, ... , \delta_T]$ from $D$;
% \State Sort  $\Delta$ to obtain  $\Delta^* = [\delta_1^*, ... , \delta_T^* ]$;
% %\STATE Total parts $\mathcal{L} = 1 + (R - 1)(1 - \tau)$;
% \For{each $r = 1,2,3,...,R $}
% \State $\zeta_{left}^r = \Delta^*[\lfloor (T-1) \times  \frac{ (r-1)}{R}\rfloor] $
% \State $\zeta_{right}^r = \Delta^*[\lfloor (T-1) \times  \frac{ r}{R}\rfloor] $
% \EndFor
% \end{algorithmic}
% \end{algorithm}

\vspace{2pt} \noindent  \textbf{Optimization. } 
%how to separate interval, and the final loss function 
We train the regression tree by imposing a classification task on the leaf probabilities and a regression task on the ground-truth interval. Specifically, when the ground-truth  score difference of the input pairs $\delta$ is in $i$-th group, \ie $\delta \in (\zeta_{\lefta}^i , \zeta_{\righta}^i )$, the one-hot label classification $L = \{\textit{l}_r\}$ is defined by assigning 1 to the $i$-th node and the regression label is set as $\sigma_i = \frac{\delta - \zeta_{\lefta}^i}{\zeta_{\righta}^i -\zeta_{\lefta}^i}$. 

For each video pair in the training data with classification label $\{l_r\}_{r=1}^R$ and regression label $\{\sigma_r\}_{r=1}^R$, the objective function for the classification task and regression task can be written as:
\begin{small}\begin{equation*}
    \begin{split}
        &J_{\rm cls} = -\sum_{r=1}^{R}\left(l_r\log(P_r) + (1 - l_r)\log(1 - P_r)\right)    \\   
&J_{\rm reg} = \sum_{r=0}^{R} \mathbb{I}(l_r = 1)(\hat{\sigma}_r - \sigma_r)^2,
    \end{split}
\end{equation*}\end{small}
where $\{P_r\}_{r=1}^R$ and $\{\hat{\sigma}_r\}_{r=1}^R$ are the predicted leaf probabilities and regression results. The final objective function for the video-pair is:
\begin{small}\begin{equation}
J = J_{\rm cls} + J_{\rm reg}.
\label{equ:final}
\end{equation}\end{small}

 \noindent  \textbf{Inference. } 
The overall regression process of the proposed group-aware regression tree can be written as:
\begin{small}\begin{equation}
    \mathcal{R}_{\Theta}(f_{(n,m)}) = \hat{\sigma}_{r^*}(\zeta_{\righta}^{r^*}-\zeta_{\lefta}^{r^*}) + \zeta_{\lefta}^{r^*},
\end{equation}\end{small}
where ${r^*}$ is the group with the highest probability. In our implementation, we also adopt a multi-exemplar voting strategy. Given an input video $v_{\rm test}$, 
we select $M$ exemplars from training data to construct $M$ pairs using these $M$ different exemplars $\{v_{\rm train}^m\}_{m=1}^M$ whose scores are $\{s_{\rm train}\}_{m=1}^M$. The process of multi-exemplar voting can be summarized as:
\begin{small}\begin{equation}
\hat{s}_{\rm test}^m = \mathcal{R}_{\Theta}(\mathcal{F}_{\mathcal{W}}(v_{\rm test},v_{\rm train}^m)) + s_{\rm train}^m,    
\end{equation}
\begin{equation}
\hat{s}_{\rm test} = \frac{1}{M} \sum_{m=1}^M \hat{s}_{\rm test}^m, m=1,2,...,M.    
\end{equation}\end{small}

\section{Experiments}

\begin{table*}[!h]
\small
\caption{Comparisons of Spearman's Correlation and R-$\ell_2$ Distance on the AQA-7 dataset. $\sharp$ indicts our implementation.} 
\label{tab:aqa-7}
% \vspace{-5pt}
\centering
\setlength{\tabcolsep}{3mm}{
\begin{tabular}[\linewidth]{l | c c c c c c | c | c}
\toprule[1.5pt]
Sp. Corr & Diving & Gym Vault & BigSki. & BigSnow. & Sync. 3m & Sync. 10m & Avg. Corr. & Year\\
\midrule[1.2pt] 
% \hline
Pose+DCT~\cite{DBLP:conf/eccv/PirsiavashVT14} & 0.5300 & 0.1000 & -- & -- & -- & -- & -- & 2014\\
ST-GCN~\cite{DBLP:conf/aaai/YanXL18} & 0.3286 & 0.5770 & 0.1681 & 0.1234 & 0.6600 & 0.6483 & 0.4433 & 2018\\
C3D-LSTM~\cite{DBLP:conf/cvpr/ParmarM17} & 0.6047 & 0.5636 & 0.4593 & 0.5029 & 0.7912 & 0.6927 & 0.6165 & 2017\\
C3D-SVR~\cite{DBLP:conf/cvpr/ParmarM17} & 0.7902 & 0.6824 & 0.5209 & 0.4006 & 0.5937 & 0.9120 & 0.6937 & 2017\\
JRG~\cite{DBLP:conf/iccv/JiaHuiaction} & 0.7630 & 0.7358 & 0.6006 & 0.5405 & 0.9013 & \textbf{0.9254} & 0.7849 & 2019 \\
I3D+MLP$^*$~\cite{musdl}  & 0.7438 & 0.7342 & 0.5190 & 0.5103 & 0.8915 & 0.8703 & 0.7472 & 2020 \\
USDL~\cite{musdl} & 0.8099 & 0.7570 & 0.6538 & \textbf{0.7109} & 0.9166 & 0.8878 & 0.8102 & 2020 \\
\hline
I3D + MLP$^{*\sharp}$ & 0.8685 & 0.6939 & 0.5391 & 0.5180 & 0.8782 & 0.8486 & 0.7601 &  \\
\rowcolor{Gray} CoRe + GART$^*$ & \textbf{0.8824} & \textbf{0.7746} & \textbf{0.7115} & 0.6624 & \textbf{0.9442} & 0.9078 & \textbf{0.8401}  & \\
\midrule[1.2pt]
R-$\ell_2$($\times$100) & Diving & Gym Vault & BigSki. & BigSnow. & Sync. 3m & Sync. 10m & Avg. R-$\ell_2$ & Year\\
\midrule[1.2pt]
C3D-SVR~\cite{DBLP:conf/cvpr/ParmarM17} & 1.53 & 3.12 & 6.79 & 7.03 & 17.84 & 4.83 & 6.86 & 2017\\
USDL~\cite{musdl} & 0.79 & 2.09 & 4.82 & 4.94 & 0.65 & \textbf{2.14} & 2.57 & 2020 \\
\hline
I3D + MLP$^{*\sharp}$  & 0.81 & 2.54 & 6.06 & 5.31 & 1.41 & 3.08 & 3.20 &  \\
\rowcolor{Gray} CoRe + GART$^*$  & \textbf{0.64} & \textbf{1.78} & \textbf{3.67}& \textbf{3.87} & \textbf{0.41} & 2.35  & \textbf{2.12} &  \\
\bottomrule[1.5pt]
\end{tabular}}
% \vspace{-10pt}
\end{table*}

\subsection{Datasets and Experiment Settings} \label{sec:setup}
 \noindent \textbf{Datasets.} We perform experiments on three widely used AQA benchmarks including AQA-7~\cite{DBLP:conf/wacv/ParmarM19}, MTL-AQA~\cite{DBLP:journals/corr/abs-1904-04346} and JIGSAWS~\cite{gao2014jhu}. For more details about the datasets, please refer to the Supplementary.

\vspace{2pt} \noindent \textbf{Evaluation Protocols.}
In order to compare with the previous work~\cite{DBLP:conf/iccv/JiaHuiaction,DBLP:conf/wacv/ParmarM19,DBLP:journals/corr/abs-1904-04346,musdl} in AQA, we adopt Spearman's rank correlation as an evaluation metric. Spearman's correlation is defined as:
\begin{equation}
\rho = \frac {\sum_i(p_i-p)(q_i-q)}{\sqrt{\sum_i(p_i-\bar p)^2\sum_i(q_i-\bar q)^2}},
\end{equation}
were $p$ and $q$ represent the ranking for each sample of two series respectively. We also follow the previous work to use Fisher's z-value~\cite{DBLP:conf/wacv/ParmarM19} when measure the average performance across actions.

We also propose a stricter metric to measure the performance of AQA models more precisely, called relative L2-distance (R-$\ell_2$). 
Given the highest and lowest scores for an action $s_{\max}$ and $s_{\min}$, R-$\ell_2$ is defined as:
\begin{equation}
\text{R-}\ell_2(\theta) = \frac{1}{K} \sum^K_{k=1} \left(\frac{|s_k - \hat{s}_k|}{s_{\max} - s_{\min}}\right)^2,
\end{equation}
where $s_k$ and $\hat{s}_k$ represent ground-truth score and prediction for $k$-th sample, respectively. We use R-$\ell_2$ instead of traditional L2-distance because different actions have different scoring intervals. Comparing and averaging $\ell_2$ distance among different classes of actions is meaningless and confusing. Our proposed R-$\ell_2$ is different from Spearman's correlation: Spearman's correlation focuses more on the ranks of the predict scores while our R-$\ell_2$ focuses on the numerical values.

% $\theta$ is a tolerance threshold. If error between prediction and ground-truth is less than the threshold, the error will be ignored. $K$ is the size of dataset. In the following experiments, we report the results with $\theta=0$. Results with different $\theta$ can be found in Supplementary Material.

% Compared to previous metrics like Spearman's correlation, the proposed  R-$\ell_2$ metric has two key advantages: 1) our metric can judge a single prediction while Spearman's correlation requires the whole test set, which makes our metric more flexible; 2) our metric is stricter and more reasonable especially when the test set is relatively small. For example, diver A and diver B get score of 95 and 65 respectively by human professional judges. If the predictions of these two actions are 80 and 30, it is a prefect prediction under the Spearman's correlation metric, while our metric can clearly reflect the prediction performance.

\vspace{2pt} \noindent \textbf{Implementation Details:}
% Our proposed methods were built on the Pytorch toolbox~\cite{paszke2017automatic} and 
% implemented on a system with the Intel(R) Xeon(R) CPU E5-2620 v4 @ 2.10GHz
%
% We trained our model with two TITAN XP GPUs.
%
We adopt the I3D model pre-trained on Kinetics~\cite{DBLP:conf/cvpr/CarreiraZ17} dataset as the feature extractor $\mathcal{F}_{\mathcal{W}}$. For all the experiments, we set the depth of GART to $d=5$ and the node feature dimension as 256. The initial learning rate is 1e-3 for the regression tree and 1e-4 for the I3D backbone. We use Adam~\cite{Kingma2014Adam} optimizer, and weight decay is set to zero. we select 10 exemplars for an input test video during inference and vote for the final score using the multi-exemplar voting strategy.
% backbone
In experiments on AQA-7 and MTL-AQA, we follow~\cite{musdl,DBLP:conf/iccv/JiaHuiaction,DBLP:conf/wacv/ParmarM19,DBLP:journals/corr/abs-1904-04346} to extract 103 frames for each video clip, and segment them into 10 overlapping snippets, each containing 16 continuous frames. In JIGSAWS, we follow~\cite{musdl} to evenly sampled out 160 frames to form 10 non-overlapping 16-frame snippets.
% exemplar
In AQA-7 and JIGSAWS, we select the exemplar video only according to the coarse category of the video. For example, if the input video is from \textit{single diving-10m platform} in AQA-7, we randomly select an exemplar video from the training set of \textit{single diving-10m platform} in AQA-7. In MTL-AQA dataset, since there are annotations about the degree of difficulty (DD) for diving sports, we select the exemplar based on both the category and the degree of difficulty. Note that this implementation is consistent with the real-world scenario since DD is known to all judges before the action is completed.

We report the performance of the following methods in experiments including the baseline method and different versions of our methods\footnote{We use $*$ to indicate that we did not use DD in both training and test.}:
\begin{itemize} \setlength\itemsep{0pt}
\item \textbf{I3D + MLP} and \textbf{I3D + MLP$^{*}$}(Baseline) : Most existing works adopt this strategy. We use I3D~\cite{DBLP:conf/cvpr/CarreiraZ17} to encode a single input video, and predict the score based on the feature with a 3-layer MLP. MSE loss between the prediction and the ground-truth is used to optimize the model.

\item \textbf{CoRe + MLP} and \textbf{CoRe + MLP$^*$}: We reformulate the regression problem as mentioned in Sec.~\ref{sec:p1}. We choose exemplar videos from the training set to construct the video pairs and also use MSE loss for optimization.

\item \textbf{I3D + GART} and \textbf{I3D + GART$^*$}: We replace the regression sub-network (MLP) with our group-aware regression tree in the baseline method. We use the loss defined in Equation~\eqref{equ:final}

\item \textbf{CoRe + GART} and \textbf{CoRe + GART$^*$}: The proposed method in Section.~\ref{approach}.
\end{itemize}
Note that we did not evaluate some of them on the AQA-7 and JIGSAWS datasets due to the absence of degree of difficulty annotations.

\subsection{Results on AQA-7 dataset}

\begin{figure} \small 
\centering
% \vspace{-3pt}
\includegraphics[width=\linewidth]{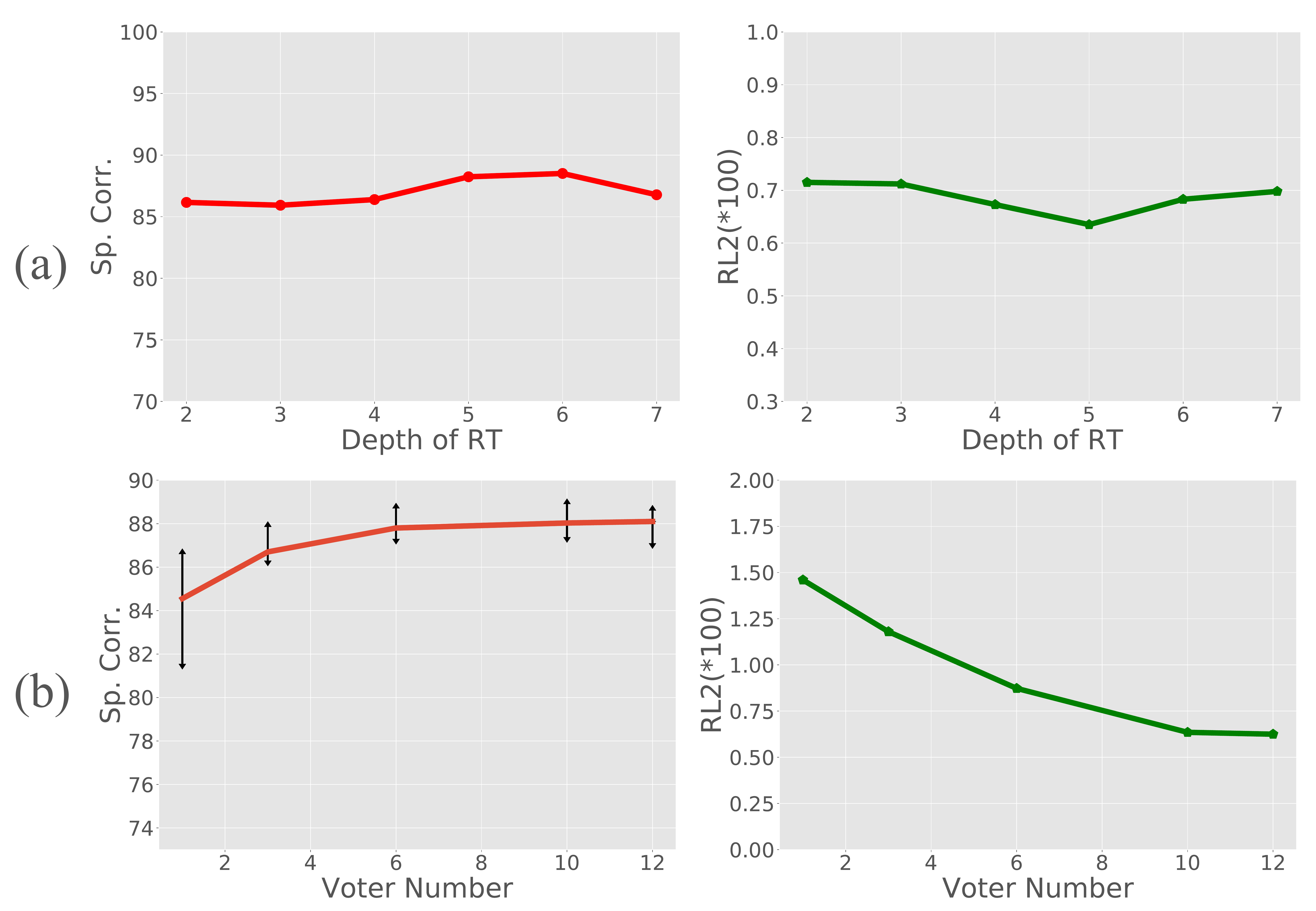}
\caption{Effects of the depth of regression tree (a) and the number of exemplars for voting (b).}\label{fig:depth}
% \vspace{-10pt}
\end{figure}

The experiment results of our method and other AQA approaches on AQA-7 are shown in Table~\ref{tab:aqa-7}. The state-of-the-art method USDL~\cite{musdl} uses Gaussian distribution to create a soft distribution label for each video, which can reduce the subjective factor from human judges on original labels. We achieve the same goal with contrastive regression. We also provide the results of the baseline I3D + MLP$^*$ on this dataset, which clearly show the performance improvement obtained by our method. We reach the best results on almost all classes in AQA-7 under both Spearman's correlation and R-$\ell_2$. Our method achieves 8.95\%, 2.32\%, 8.83\%, -6.82\%, 3.01\% and 2.25\% performance improvement for each sports class compared with USDL under Spearman's rank. Meanwhile, we achieve 0.15, 0.31, 1.15, 1.07, 0.24, -0.21 performance improvement under R-$\ell_2$. For the average correlation and average R-$\ell_2$ performance, we have nearly 3.7\% and 0.45 improvements compared to USDL model, clearly showing the effectiveness of our model.

We also conduct several analysis experiments to study the effects of the \textit{depth} of the regression tree and the vote number \textit{M} in the multi-exemplar voting on \textit{Diving} class of AQA-7 dataset.

\vspace{2pt}
\noindent \textbf{Effects of the depth of regression tree. }
In the regression tree module, the \textit{depth} of the tree is a significant hyper-parameter determining the architecture of the regression tree. We conduct several experiments on \textit{Diving} class of AQA-7 dataset with different values of \textit{depth}, ranging from 2 to 7, and set the \textit{M} to 10. As shown in Figure ~\ref{fig:depth}, our model performs better when \textit{depth} is 5 and 6, where the total number of groups is 32 and 64. However, there is a little drop in performance when \textit{depth} is smaller than 4 or bigger than 7. In general, our model is robust to different depths. % General speaking, our model is robust to this hyper-parameter.

\vspace{2pt} \noindent \textbf{Effects of the number of exemplars for voting. } The number of exemplars used in the inference phase is another important hyper-parameter. A larger number for \textit{M} means the model can refer to more exemplars while leading to larger computational cost. We conduct experiments on \textit{Diving} class to study the impact of \textit{M}. Figure~\ref{fig:depth} shows the result when the \textit{depth} of regression tree is set to 5. We observe that with $M$ increasing, the performance becomes better and the variance is lower. The improvement on Sp. Corr. becomes less significant when $M$ exceeds 10. We can also find the same trend for R-$\ell_2$.

\subsection{Results on MTL-AQA dataset}

\begin{table}
\small
\caption{Comparisons of performance with existing methods on the MTL-AQA dataset. $\sharp$ indicts our implementation.} 
\label{tab:mtl}
% \vspace{-5pt}
\centering
\begin{tabular}{l | c c | l}
\toprule[1.5pt]
Method (w/o DD) & Sp. Corr. & R-$\ell_2$($\times$100) &Year\\
\midrule[1.2pt] 
Pose+DCT~\cite{DBLP:conf/eccv/PirsiavashVT14} & 0.2682 & -- & 2014\\
C3D-SVR~\cite{DBLP:conf/cvpr/ParmarM17} & 0.7716 & -- & 2017\\
C3D-LSTM~\cite{DBLP:conf/cvpr/ParmarM17} & 0.8489 & --& 2017\\
MSCADC-STL~\cite{DBLP:journals/corr/abs-1904-04346} & 0.8472 & --& 2019\\
C3D-AVG-STL~\cite{DBLP:journals/corr/abs-1904-04346} & 0.8960 & --& 2019\\
MSCADC-MTL~\cite{DBLP:journals/corr/abs-1904-04346} & 0.8612 & --& 2019\\
C3D-AVG-MTL~\cite{DBLP:journals/corr/abs-1904-04346} & 0.9044 & --& 2019\\
I3D + MLP$^*$~\cite{musdl} & 0.8921 & 0.707 & 2020 \\
USDL~\cite{musdl}& 0.9066 & 0.654& 2020\\
MUSDL$^*$~\cite{musdl} & 0.9158 & 0.609  & 2020\\
\hline
I3D + MLP$^{*\sharp}$ & 0.9196 & 0.465 & \\
CoRe + GART$^*$  & \textbf{0.9341}   & \textbf{0.365}&  \\
%ours-RT$^*$ & ???? & \\
%ours-GC$^*$ & ???? & \\
\midrule[1.2pt]
Method (w/ DD)& Sp. Corr. & R-$\ell_2$($\times$100) &Year\\
\midrule[1.2pt] 
USDL$_{DD}$~\cite{musdl} & 0.9231 & 0.468  & 2020\\
MUSDL~\cite{musdl} & 0.9273 & 0.451  & 2020\\
\hline
I3D + MLP & 0.9381 & 0.394& \\
% Ours-CoRe+Tree$\dag$ & 0.9481 & 0.312& \\
\rowcolor{Gray} CoRe + GART &   \textbf{0.9512} & \textbf{0.260}& \\
%ours-RT & ???? & \\
%ours-GC & 0.9316 & \\
\bottomrule[1.5pt]
\end{tabular}
%   \vspace{-5pt}
\end{table}

\begin{table}
% \footnotesize
%  \centering
  \caption{Ablation study on MTL-AQA dataset}
  \label{tab:ablation} \small
%   \vspace{-5pt}
  \centering
  \begin{tabular}{l |c| c c}
  \toprule[1.5pt]
  Method  & Ablation & Sp. Corr. & R-$\ell_2$($\times$100)\\
  \midrule[1.2pt] 
%   I3D+MLP$^*$  &  & & & 0.8921& 0.707\\
%   ours-RT$^*$   &  &  &   \checkmark   & 0.9037 & 0.658 \\
%   Ours-CoRe+MLP$^*$  &  & \checkmark &      & 0.9126 & 0.489 \\
%   Ours-CoRe+Tree$^*$ &    &  \checkmark &   \checkmark  & \textbf{0.9223} & \textbf{0.438} \\
%   \midrule[1.2pt] 
%   I3D + MLP  &  &   & 0.9189 & 0.512\\
%   I3D + GART    &  &  \checkmark & 0.9281 & 0.421  \\
%   CoRe + MLP & \checkmark &     & 0.9316 & 0.381 \\
%   CoRe + GART & \checkmark & \checkmark  & \textbf{0.9512} & \textbf{0.260} \\
   I3D + MLP  & Baseline   & 0.9381 & 0.394\\
   I3D + GART & + GART & 0.9403 & 0.366\\
   CoRe + GART & + CoRe  & \textbf{0.9512} & \textbf{0.260} \\
  \bottomrule[1.5pt]
  \end{tabular}
    % \vspace{-10pt}
\end{table}

\begin{figure}[t]
  \centering
  \includegraphics[width=0.9\linewidth]{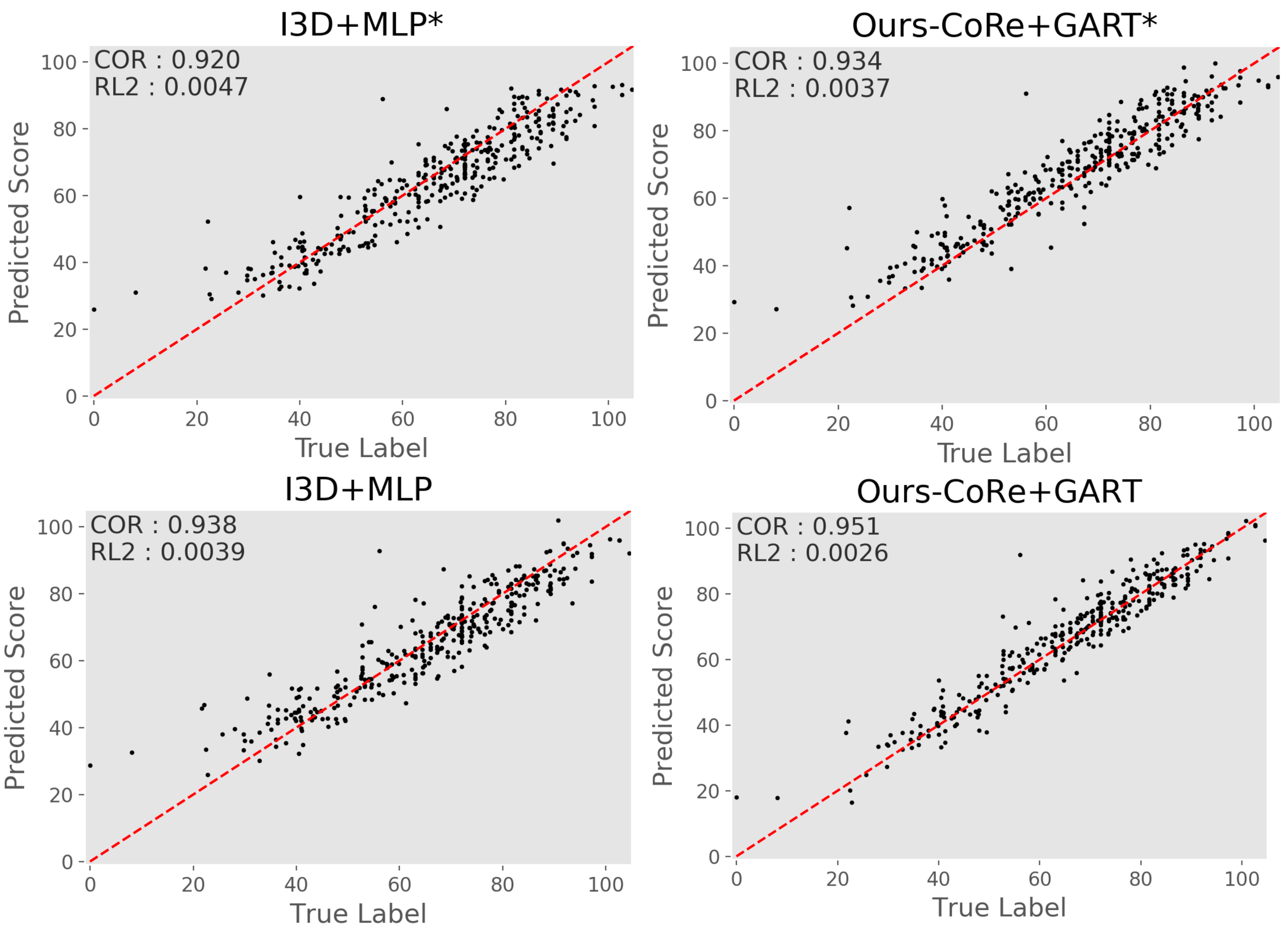}
  \caption{A comparison of different methods in scatter plot. Each point in the figure represents a video in the test set. The red line indicates the prefect predictions.
  }
  \label{fig:scatter}
% \vspace{-5pt} 
\end{figure}

\begin{figure}
  \centering
  \includegraphics[width=0.7\linewidth]{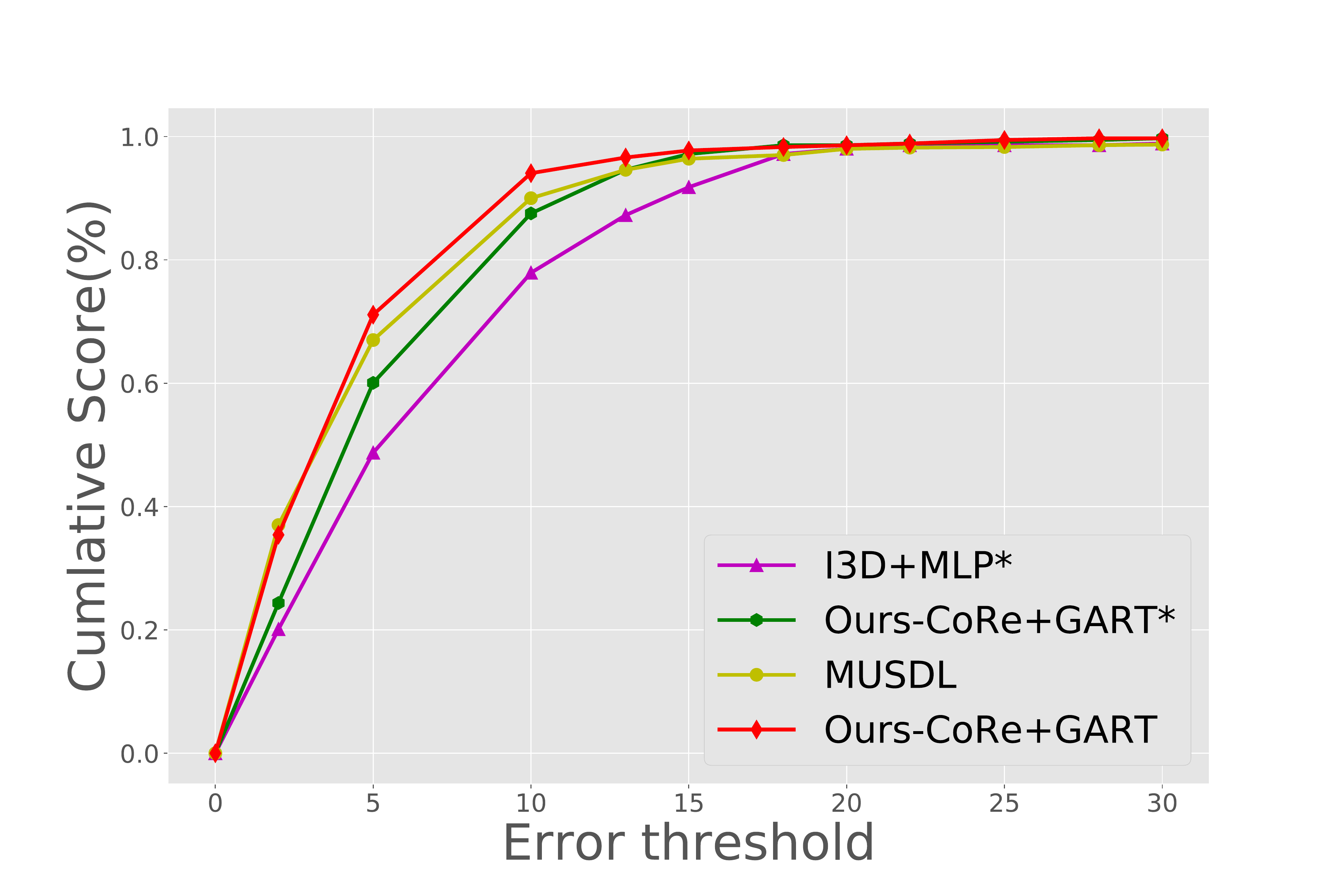}
  \caption{Cumulative score curve on MTL-AQA dataset.  The larger the area under the curve indicates the better performance.
}
  \label{fig:cumlative}
    \vspace{-5pt}
\end{figure}

\begin{figure*}[!h]
  \centering
  \includegraphics[width=0.95\linewidth]{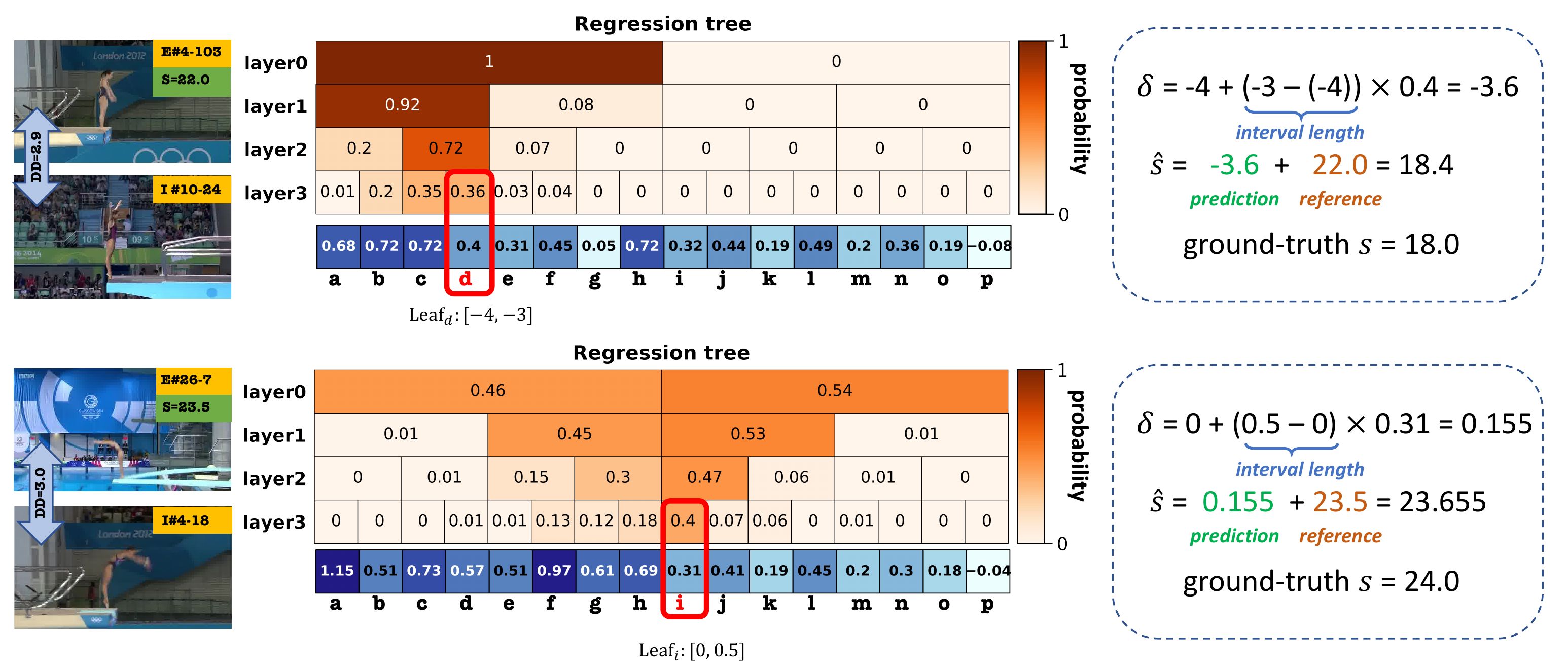} \vspace{-5pt}
  \caption{Case study. 
 The videos marked with $E$ and $I$ in the upper left corner are the exemplar and the input video, respectively. Each pair of exemplar and input videos have the same degree of difficulty (DD). We show the probability output for each layer of the regression tree and the regression value for each leaf on the right. We take the regression value of the leaf node with the highest probability as the final regression result. The very small errors between our prediction results with ground-truths demonstrate the effectiveness of our method. }
  \label{fig:sampler_vis} \vspace{-10pt}
\end{figure*}

Table~\ref{tab:mtl} shows the performance of existing methods and our method on MTL-AQA dataset. Since the degree of difficulty (DD) annotations are available for diving actions in MTL-AQA, we also verify the effects of DD on this dataset. We divide all methods into two types: some use the DD labels in the training phase (bottom part of the table) and the others (upper part of the table) do not. We see CoRe + GART$^*$ achieves respectively 2.0\% and 0.244 improvement compared to MUSDL$^*$~\cite{musdl} under Spearman's rank and R-$\ell_2$ metric without DD labels. By training with the degree of difficulty, our method becomes even better, achieving 2.6\% and 0.191 improvements compared to MUSDL under the two metrics. We conjecture that there are two reasons: one is that we can select more suitable exemplars, the other reason is that our method can exploit more information about the action from the degree of difficulty. % We conduct an additional experiment to further explore which is the main reason. In CoRe+Tree$\dag$, we did not use the degree of difficulty to select exemplars, which cause a 0.31\% and 0.062 drop in performance only, which shows our method can exploit the information from the degree of difficulty better than MUSDL. This is very useful because the degree of difficulty is easy to know in practice because it is always the case that athletes choose an action in advance, then they try to perform the chosen action. 
To have an intuitive understanding of the differences between our method and baseline methods, we visualize the prediction results in form of a scatter plot in Figure~\ref{fig:scatter}. We see our method is much more accurate compared to the baseline. By using the degree of difficulty information, the performance of our method can be further improved, where almost all the points are near the red line in the middle of the picture. In Figure~\ref{fig:cumlative}, we show the cumulative score curves of our methods and SOTA method MUSDL~\cite{musdl}. Given the error threshold $\epsilon$,  the samples whose absolute differences between their prediction and ground-truth are less than $\epsilon$ will be regarded as positive samples. It can be observed that under any error threshold, CoRe + GART (red line) shows a stronger ability to predict accurate scores.

\vspace{2pt} \noindent \textbf{Ablation Study.} We further conduct an ablation study for our method. The results are shown in Table~\ref{tab:ablation}. Comparing I3D + MLP and I3D + GART, we see when replacing MLP with our group-aware regression tree, the performance is improved by 0.0022 and 0.028 under Spearman's rank metric and R-$\ell_2$ metric, which demonstrates the effectiveness of the designs of GART. The performance is further improved when replace the I3D baseline with our proposed CoRe framework. The above results demonstrate the effectiveness of the two components of our method.

% The regression tree is more suitable for our contrastive regression framework, for every decision made in the tree node is based on the comparison. 

% \paragraph{Scatter plot} To have an intuitive understanding of the differences between our method and baseline methods, we visualize the prediction result in form of scatter plot in Figure.~\ref{fig:scatter}. The first row shows the baseline method and the proposed approach under the no-difficulty-degree setting. Our method can more accurately predict scores compared with the baseline method. In the second row, the degree of difficulty is used in the inference phase. We can clearly see that our method performs very good, almost all  points is beside the red line in the middle of picture

\begin{table}
\small
\caption{Comparisons of performance with existing methods on the JIGSAWS dataset.} 
\label{tab:jigsaws}
\centering
\begin{tabular}{l | c c c | c }
\toprule[1.5pt]
 Sp. Corr.& S & NP & KT & Avg. Corr.\\
\midrule[1.2pt] 
ST-GCN~\cite{DBLP:conf/aaai/YanXL18}& 0.31 & 0.39 & 0.58 & 0.43  \\
TSN~\cite{DBLP:conf/cvpr/ParmarM17} & 0.34 & 0.23 & 0.72 & 0.46 \\
JRG~\cite{DBLP:conf/iccv/JiaHuiaction} & 0.36 & 0.54 & 0.75 & 0.57 \\
USDL~\cite{musdl} & 0.64 & 0.63 & 0.61 & 0.63  \\
MUSDL~\cite{musdl} & 0.71 & 0.69 & 0.71 & 0.70 \\
\hline
I3D + MLP$^*$ & 0.61 & 0.68 & 0.66 & 0.65  \\
\rowcolor{Gray} CoRe + GART$^*$ &\textbf{0.84} & \textbf{0.86} & \textbf{0.86} & \textbf{0.85} \\
\toprule[1.5pt]
 R-$\ell_2$($\times$100) & S & NP & KT & Avg. \\
\midrule[1.2pt] 
I3D + MLP$^*$ & 4.795 & 11.225 & 6.120 & 7.373  \\
\rowcolor{Gray} CoRe + GART$^*$ &\textbf{5.055} & \textbf{5.688} & \textbf{2.927} & \textbf{4.556} \\
\bottomrule[1.5pt]
\end{tabular}
%   \vspace{-5pt}
\end{table}

\begin{figure}[t]
    \centering
    \includegraphics[width=\linewidth]{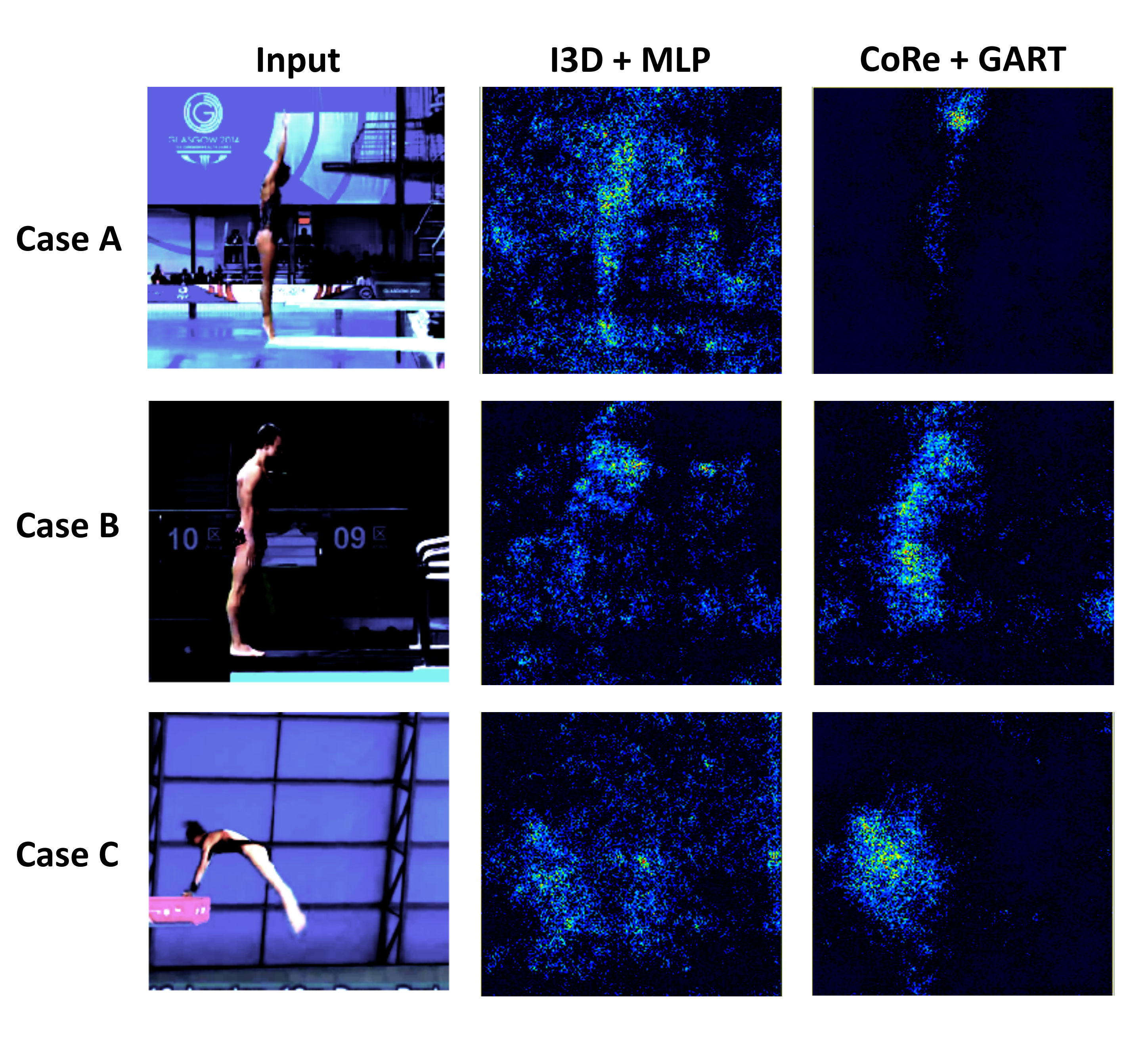}
    \caption{Visualization. We show the visualization result on MTL-AQA using Grad-CAM~\cite{selvaraju2017grad}. Our method can focus on the regions that are critical to assess the action quality.}
    \label{fig:viz}
    \vspace{-15pt}
\end{figure}
\vspace{2pt} \noindent \textbf{Case Study. }
In order to have a deeper understanding of the behavior of our model, we present a case study in Figure~\ref{fig:sampler_vis}.  Based on the comparison between input and exemplar, the regression tree determines the relative score from coarse to fine. The first layer of the regression tree tries to determine which video is better, and the following layers try to make the prediction more accurate.  The first case in the figure shows the behavior when the difference between input and exemplar is large, and the second case shows the behavior when the difference is small. In both situations, our model can give satisfactory predictions.

\subsection{Results on JIGSAWS}
We also conduct experiments on this surgical action dataset JIGSAWS.
Four-fold cross-validation is used following previous works~\cite{musdl,DBLP:conf/iccv/JiaHuiaction}.
Table~\ref{tab:jigsaws} shows the experiment results. CoRe + GART$^*$ largely improves the previous state-of-the-arts. Our method also obtains a more balanced performance in different action classes.

\subsection{Visualization} To further prove the effectiveness of our method, we visualize the baseline model (I3D + MLP) and our best model (CoRe + GART) using Grad-CAM~\cite{selvaraju2017grad} on MTL-AQA, as is shown in Figure~\ref{fig:viz}. We observe that our method can focus on certain regions (hands, body, \etc), which indicates our contrastive regression framework can alleviate the influence caused by the background and pay more attention to the discriminative parts.

\section{Conclusions}
In this paper, we have proposed the \textit{CoRe} framework for action quality assessment, which learns the relative scores based on the exemplars.  We have also devised a group-aware regression tree to convert the conventional score regression into a coarse-to-fine classification task and a regression task in small intervals. The experiments on three AQA datasets have demonstrated the effectiveness of our approach. We expect the introduction of \textit{CoRe} provides a new and generic solution for various AQA tasks.

\subsection*{Acknowledgements}
This work was supported in part by the National Natural Science Foundation of China under Grant U1813218, Grant U1713214, and Grant 61822603, in part by a grant from the Beijing Academy of Artificial Intelligence (BAAI), and in part by a grant from the Institute for Guo Qiang, Tsinghua University.

\begin{appendix}

\section{Datasets}
We now describe the datasets we used in our experiments in detail.

\vspace{5pt}
\noindent\textbf{AQA-7}~\cite{DBLP:conf/wacv/ParmarM19}: 
AQA-7 contains 1,189 samples from seven different actions collected from winter and summer Olympic Games. It contains two dataset released before: UNLV-Dive \cite{DBLP:conf/cvpr/ParmarM17} is named \textit{single diving-10m platform} in AQA-7, contains 370 samples. UNLV-Vault \cite{DBLP:conf/cvpr/ParmarM17} is named \textit{gymnastic vault} in AQA-7, contains 176 samples; The other action classes are newly collected in this dataset: \textit{synchronous diving-3m springboard} contains 88 samples and \textit{synchronous diving-10m platform} contains 91 samples. \textit{big air skiing} cantains 175 samples and \textit{big air snowboarding} contains 206 samples.
 
\vspace{5pt}
\noindent\textbf{MTL-AQA}~\cite{DBLP:journals/corr/abs-1904-04346}:
The MTL-AQA dataset contains all kinds of \textit{diving} actions, which is the largest AQA dataset up to date. There are 1,412 samples collected from 16 difference world events. 
% including both male and female, both individual and synchronous divers,
% both 3m springboard and 10m platform and different views.
The annotations in this dataset are various, including the degree of difficulty (DD), scores from each judge (totally 7 judges), type of diver's action, and the final score. We adopt the evaluation protocol suggested in \cite{DBLP:journals/corr/abs-1904-04346} in our experiments.
% Note that, the final score of a diving action comes from two parts: difficulty level and quality of execution %~\footnote{Summarizing scores from different judges:the top-two and bottom-two judges' scores will be eliminated. quality of execution is the sum of the remaining three judges' scores}. 
% The final score is the product of them, and difficulty level is known to judges before the action starts.
% Evaluation protocol suggested in \cite{DBLP:journals/corr/abs-1904-04346} was followed in our experiments.

\vspace{5pt}
\noindent\textbf{JIGSAWS}~\cite{gao2014jhu}: JIGSAWS is a surgical actions dataset containing 3 type of surgical task: "\textit{Suture}(S)", "\textit{NeedlePassing}(NP)" and "\textit{Knotted}(KT)".
For each task, each video sample is annotated with multiple annotation scores assessing different aspects of surgical actions, and the final
score is the sum of those sub-scores. 
% Every action in the dataset is recorded by the left and right cameras at the same time, but the difference between two views is very small. 
% So we only trained and tested from one perspective.
We adopt a similar four-fold cross validation strategy as~\cite{DBLP:conf/iccv/JiaHuiaction, musdl}.

\section{More Discussions}
\begin{figure}[h]
  \centering
  \includegraphics[width=\linewidth]{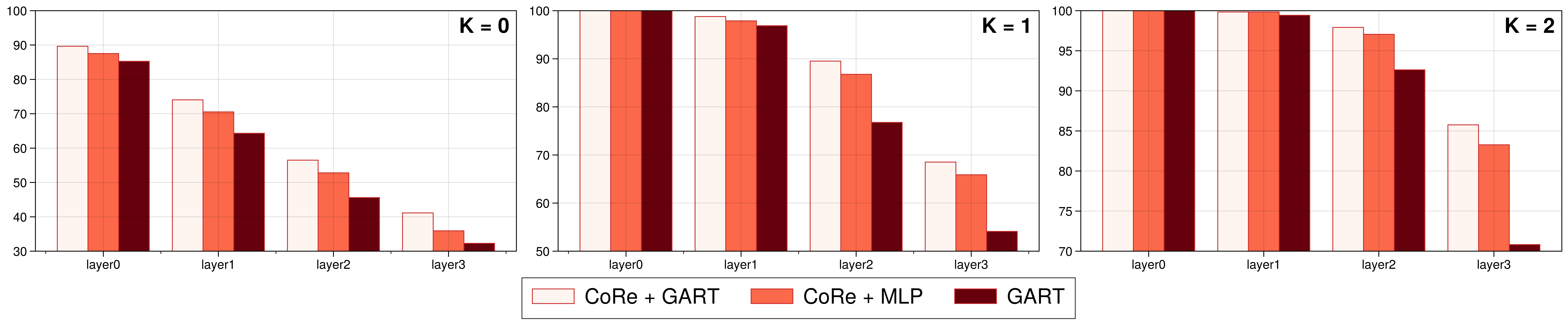} 
  \caption{Classification accuracy for each layer of the group-aware regression tree. \textit{CoRe + GART} is our final method, combining contrastive regression and group-aware regression tree together.  \textit{CoRe + MLP} uses an MLP to replace the regression tree and the \textit{GART} method only keeps the regression tree without using the contrastive regression framework. $K$ is a tolerance threshold, which indicates classifying a pair into the nearest-K groups is still regarded as a correct classification. }
  \label{fig:layer_acc}
\end{figure}

\paragraph{More analysis on the regression tree. }
 To better understand the prediction process of the regression tree, we also investigate the prediction accuracy of each layer in the regression tree on the MTL-AQA dataset, as shown in Figure~\ref{fig:layer_acc}. We also compare the results with two baseline methods.  Comparing \textit{CoRe + GART} and \textit{GART}, we can see \textit{CoRe + GART} performs better in each layer under all values of $K$, which indicates measuring relative score between input and exemplar is more effective than predicting the final score directly. Comparing two CoRe-based methods, we see the group-aware regression tree measures relative score more accurately.  % ~\footnote{The leaves in our regression tree represent a certain interval. Higher classification accuracy leads to a smaller error between the final prediction and the ground-truth.}
 
\begin{figure*}[h]
   \centering
   \includegraphics[width=\linewidth]{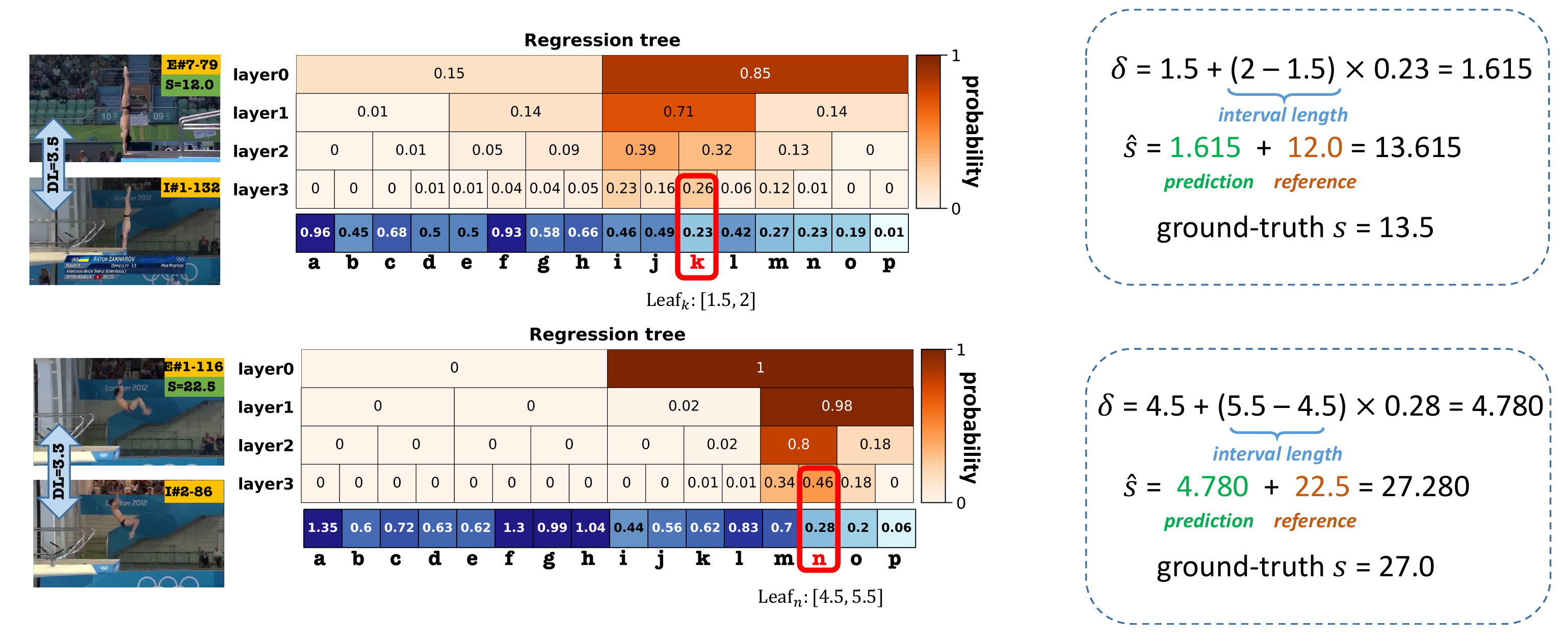}
  \caption{\small Case study. The videos marked with $E$ and $I$ in the upper left corner are the exemplar and the input video, respectively. Each pair of exemplar and input videos have the same degree of difficulty (DD). We show the probability output for each layer of the regression tree and the regression value for each leaf on the right. We take the regression value of the leaf node with the highest probability as the final regression result.}
   \label{fig:sampler_vis}
 \end{figure*}

\paragraph{More analysis on CoRe. } Another advantage of our proposed CoRe is that CoRe could alleviate the subjectiveness from human judges by predicting the difference, despite the fact that the exemplar video is also annotated by human judges. Formally, we can assume a score $\mathbf{x} = x + n$ can be decompose as the actual value $x$ and a subjectiveness term $n$ that subjects to normal distribution $\mathcal{N}(0, \sigma^2)$. If we directly predict $\mathbf{x}$, the variance of subjectiveness term is $\sigma^2$. By introducing $M$ exemplar videos with scores $\{\mathbf{x}_1,...,\mathbf{x}_M\}$,  our goal is to predict the difference 
\begin{equation}
    \delta = \frac{1}{M} \sum_i (\mathbf{x} - \mathbf{x}_i),
\end{equation}
which also subjects to a normal distribution:
\begin{equation}
\delta\sim \mathcal{N}\left(\frac{1}{M} \sum_i (x - x_i), \frac{2}{M}\sigma^2\right)
\end{equation}
We see the prediction becomes closer to the actual value when $M>2$. The empirical results in Figure 5(b) in the original paper also support our assumption. 

\paragraph{More analysis on R-$\ell_2$. } To more precisely measure the AQA performance, we propose a stricter metric, called relative L2-distance (R-$\ell_2$), to measure the performance of the score prediction model. We use R-$\ell_2$ instead of traditional L2-distance because different actions may have different scoring intervals. Comparing and averaging $\ell_2$ distance among different classes of actions is may be confusing in some cases.
Given the highest and lowest scores for an action $s_{max}$ and $s_{min}$, R-$\ell_2$ is defined as:
\begin{equation}\small
\text{R-}\ell_2(\theta) = \frac{1}{K} \sum^K_{k=1} (\frac{ \max(|s_k - \hat{s}_k|- \theta, 0)  }{s_{max} - s_{min}})^2.
\end{equation}
$s_k$ and $\hat{s}_k$ represent ground-truth score and prediction for $k^{th}$ sample. $\theta$ is a tolerance threshold. If error between prediction and ground-truth is less than the threshold, the error will be ignored. $K$ is the size of dataset. 

Compared to previous metrics like Spearman's correlation, the proposed  R-$\ell_2$ metric has two key advantages: 1) our metric can judge a single prediction while Spearman's correlation requires the whole test set, which makes our metric more flexible; 2) our metric is stricter and more reasonable especially  when the test set is relatively small. For example, diver A and diver B get score of 95 and 65 respectively by human professional judges. If the predictions of these two actions are 80 and 30, it is a prefect prediction under the Spearman's correlation metric, while our metric can clearly reflect the prediction performance.

\section{Case study}
We conduct two more case studies here, as shown in Figure~\ref{fig:sampler_vis}. 
Based on the comparison between the input and the exemplar, the regression tree determines the relative score from coarse to fine. The first layer of the regression tree tries to determine which video is better, and the following layers try to make this determination more accurate.  The first case in the figure shows the behavior when the difference between the pair is small, while the second case shows the behavior when this difference is large. When the difference between the two videos is large, it is easy to make the prediction. While the difference is small, the classification task is more difficult, but our method can still give a relatively accurate judgment. We see the proposed contrastive regression framework and the regression tree are two key techniques to achieve accurate score prediction.

\end{appendix}

{\small
\bibliographystyle{ieee_fullname}
\bibliography{ref}
}
\end{document}